\documentclass[runningheads]{llncs}
\usepackage{graphicx}
\begin{document}
\title{Learned super resolution ultrasound for improved breast lesion characterization}
\titlerunning{Learned super resolution US for improved breast lesion characterization}
%
%
\author{Bar-Shira, Or\inst{1} \and
Grubstein, Ahuva\inst{2,3} \and
Rapson, Yael\inst{2,3} \and
Suhami, Dror\inst{2,3} \and
Atar, Eli\inst{2,3} \and
Peri-Hanania, Keren\inst{1} \and
Rosen, Ronnie\inst{1} \and
Eldar, Yonina C.\inst{1}}
%
\authorrunning{O. Bar-Shira et al.}
%
\institute{Department of Computer Science and Applied Mathematics, Weizmann Institute of Science, Israel \and
Radiology Department, Beilinson Campus, Rabin Medical Center, Israel \and
Sackler Faculty of Medicine, Tel Aviv University, Israel}
\maketitle              
\begin{abstract}
Breast cancer is the most common malignancy in women. Mammographic findings such as microcalcifications and masses, as well as morphologic features of masses in sonographic scans, are the main diagnostic targets for tumor detection. However, improved specificity of these imaging modalities is required. A leading alternative target is neoangiogenesis. When pathological, it contributes to the development of numerous types of tumors, and the formation of metastases. Hence, demonstrating neoangiogenesis by visualization of the microvasculature may be of great importance. Super resolution ultrasound localization microscopy enables imaging of the microvasculature at the capillary level. Yet, challenges such as long reconstruction time, dependency on prior knowledge of the system Point Spread Function (PSF), and separability of the Ultrasound Contrast Agents (UCAs), need to be addressed for translation of super-resolution US into the clinic. In this work we use a deep neural network architecture that makes effective use of signal structure to address these challenges. We present in vivo human results of three different breast lesions acquired with a clinical US scanner. By leveraging our trained network, the microvasculature structure is recovered in a short time, without prior PSF knowledge, and without requiring separability of the UCAs. Each of the recoveries exhibits a different structure that corresponds with the known histological structure. This study demonstrates the feasibility of in vivo human super resolution, based on a clinical scanner, to increase US specificity for different breast lesions and promotes the use of US in the diagnosis of breast pathologies.
\end{abstract}
\section{Introduction}
Breast cancer is the most commonly occurring cancer in women and the second most common cancer overall. Breast cancer diagnosis in its early stages plays a critical role in patient survival. Mammographic findings such as microcalcifications and masses, as well as morphologic features of masses in sonographic scans, are the main diagnostic targets for tumor detection. However, there is a continued need to improve the sensitivity and specificity of these imaging modalities.  A leading alternative traget is neoangiogenesis. Neoangiogenesis is a process of development and growth of new capillary blood vessels from pre-existing vessels. When pathological, it contributes to the development of numerous types of tumors, and the formation of metastases~\cite{goussia2018associations,fox2007breast,toi1995tumor}. Robust, precise, fast, and cost-effective in-vivo microvascular imaging can demonstrate the impaired or remodeled microvasculature, thus, it may be of great importance for early detection and clinical management of breast pathologies~\cite{van2020super}.

Diagnostic imaging plays a critical role in healthcare, serving as a fundamental asset for timely diagnosis, disease staging, and management as well as for treatment strategy and follow-up. Among the diagnostic imaging options, US imaging~\cite{cosgrove2010imaging} is uniquely positioned, being a highly cost-effective modality that offers the clinician and the radiologist a high level of interaction, enabled by its real-time nature and portability~\cite{van2019deep}. The conventional US is limited in resolution by diffraction, hence, it does not resolve the microvascular architecture. Using encapsulated gas microbubbles with size similar to red blood cells as UCAs, extends the imaging capabilities of US, allowing imaging of fine vessels with low flow velocities. Specifically, Contrast-Enhanced US (CEUS) enables real-time hemodynamic and noninvasive perfusion measurements with high-penetration depth. However, as the spatial resolution of conventional CEUS imaging is bounded by diffraction, US measurements are still limited in their capability to resolve the microvasculature~\cite{bar2018sushi}.

The diffraction limited spatial resolution was recently surpassed with the introduction of super resolution Ultrasound Localization Microscopy (ULM). This technology facilitated fine visualization and detailed assessment of capillary blood vessels. ULM relies on concepts borrowed from super-resolution fluorescence microscopy techniques such as Photo-Activated Localization Microscopy (PALM) and Stochastic Optical Reconstruction Microscopy (STORM)~\cite{betzig2006imaging,rust2006sub}, which localize individual fluorescing molecules with subpixel precision over many frames and sum all localizations to produce a super-resolved image. In the ultrasonic version, CEUS is used~\cite{couture2009ultrafast,couture2012ultrasound}, where the fluorescent beacons are replaced with UCAs which are scanned with an ultrasonic scanner. When the concentration of the UCAs is sparse, individual UCAs in each diffraction-limited ultrasound frame are resolvable. Thus, when the system PSF is known, localization with micrometric precision can be obtained for each UCA. As these contrast agents flow solely inside the blood vessels, the accumulation of these subwavelength localizations facilitates the recovery of a super-resolved map of the microvasculature~\cite{bar2018sushi,christensen2020super}.

Various works based on the above idea were illustrated in vitro, and in vivo on different animal models~\cite{christensen2020super}. However, most of the super resolution US implementations to date are still limited to research ultrasound scanners with high frame-rate (HFR) imaging capability~\cite{huang2020super}. First in vivo human demonstrations using clinical scanners with low imaging frame-rates ($<15$ Hz) were shown for breast cancer~\cite{opacic2018motion,dencks2018clinical}, lower limb~\cite{harput2018two}, and prostate cancer~\cite{kanoulas2019super}. However, all methods rely on prior parameters calibration that characterize the system PSF to facilitate accurate identification of UCA signals. Further in human demonstrations were recently achieved for different internal organs and tumors~\cite{huang2020super}. Nevertheless, the processing was performed on in-phase/quadrature (IQ) data that was acquired with a high frame rate US scanner; both are not commonly available in clinical practice.

While super resolution ULM avoids the trade-off between resolution and penetration depth, it gives rise to a new trade-off that balances localization precision, UCA concentration, acquisition time, reconstruction time, and dependency on prior knowledge such as the PSF of the system. These challenges need to be addressed for translation of super-resolution US into the clinic where high UCAs concentrations, limited time, significant organ motion and lower frame-rate imaging are common~\cite{van2019deep}.

In this work, we suggest a new approach to enable increased specificity in characterization of breast lesions by relying on a model-based convolutional neural network called deep unrolled ULM, suggested by van Sloun et al.~\cite{van2019deep}. Although the method was used before for an in vivo animal model with a high frame rate scanner, here it is used for the first time for in vivo human scans with a clinical US scanner operating at low frame rates. The network makes effective use of structural signal priors to perform localization microscopy in dense scenarios. Furthermore, no prior knowledge about the system PSF is required at inference. A learned PSF alleviates the dependency on the user experience thus making the process of super-resolution more accessible. We present preliminary in vivo human results on a standard clinical US scanner for three lesions in breasts of three patients. The results demonstrate a 31.25 µm spatial resolution. The three recoveries exhibit three different vasculature patterns, one for each lesion, that correspond with the histological known structure. To the best of our knowledge, this is the first in vivo human super resolution imaging leveraging deep learning using a standard clinical scanner with low frame rates to help differentiate between breast lesions. This study demonstrates the feasibility of a learning-based approach for in human super resolution, based on a clinical scanner, to increase the specificity of US for characterization of different lesion types and promotes the use of US in the diagnosis of breast pathologies.

\section{Materials and Methods}
\subsection{Clinical Measurements}
The clinical CEUS data were acquired at the Department of Radiology, Rabin Medical Center, Petah Tikva, Israel. The study was approved by the Helsinki comitee of Rabin Medical Center, under number 0085-19-RMC. Written informed consent was obtained from all participants for CEUS imaging and the use of data for the study of improving US breast imaging. Twenty-one women aged 35-64 years with breast lesions were enrolled into this study. Measurement data of three different patients having three types of breast lesions were retrospectively evaluated. A fibroadenoma- a benign solid mass of the breast, A cyst- a benign fluid filled lesion, and an invasive ductal carcinoma- a malignant mass.

For the measurements, the patient was lying supine in a stable position. Each patient was intravenously administered 5 mL of contrast material containing 40 $\mu$L of sulphur hexafluoride microbubbles (SonoVue, Bracco, Milan, Italy), followed by a 5 mL saline flush. Real-time B-mode was used to guide the image plane and real time CEUS was used to monitor the UCAs signal right after the injection. The B-mode and CEUS images were saved for post-processing offline. During data acquisition, the patients and the dedicated breast radiologist were given oral and written instructions to be exceedingly stable to reduce out-of-plane motion.

\subsection{Ultrasound Imaging Settings}
The CEUS measurements were performed in a contrast specific mode to enhance sensitivity to UCAs while suppressing backscattering from tissue. A SL10-2 linear transducer (bandwidth 2–10 MHz) connected to a Hologic SuperSonic Mach 30 (SuperSonic Imagine, Aix-en-Provence, France) was used. The mechanical index during the examinations was 0.07. Both the B-mode images and contrast mode images were recorded with frame rate of 25 Hz; 6286 frames were recorded for each measurement (about 4 minutes).

\subsection{Image Preprocessing}
All the following procedures were implemented in MATLAB 2019a (Mathworks, Natick, MA, USA).

The total number of detected vessels within the acquisition time is influenced by the flow-rate of UCAs in the vessels which depends on the blood flow in the vessels and on the UCA concentration in the blood~\cite{dencks2018clinical}. To increase the number of blood vessels detected we looked at the time intensity curve (TIC) calculated as the mean intensity at each frame from the CEUS sequence and viewed frames after the maximum intensity was reached. This was used as an indication that the UCAs are in wash out phase.

To reduce out-of-plane motion, which can severely hinder recovery of the vasculature~\cite{dencks2018clinical}, we divided the measurements into subsequences of similar frames. This was achieved by computing the cross-correlation of the B-mode images across a manually selected region of interest (ROI) with sufficient contrast. Consecutive frames were assigned to the same subsequence if their cross-correlation was above 90\%. We considered subsequences containing above 1000 frames. Small motions were corrected for the selected subsequences using image registration that accounts for translation. The transformation matrix was computed using the B-mode images, where the first frame of a sequence was the reference frame. Because the UCAs were not visible in the B-mode sequences, the motion estimation was not disturbed. A spatiotemporal singular-value-decomposition (SVD) based filter was applied to the CEUS frames to extract moving UCAs signals, which were then used for the super resolution recovery.

\subsection{Localization of UCAs}
\begin{figure}
\includegraphics[width=\textwidth]{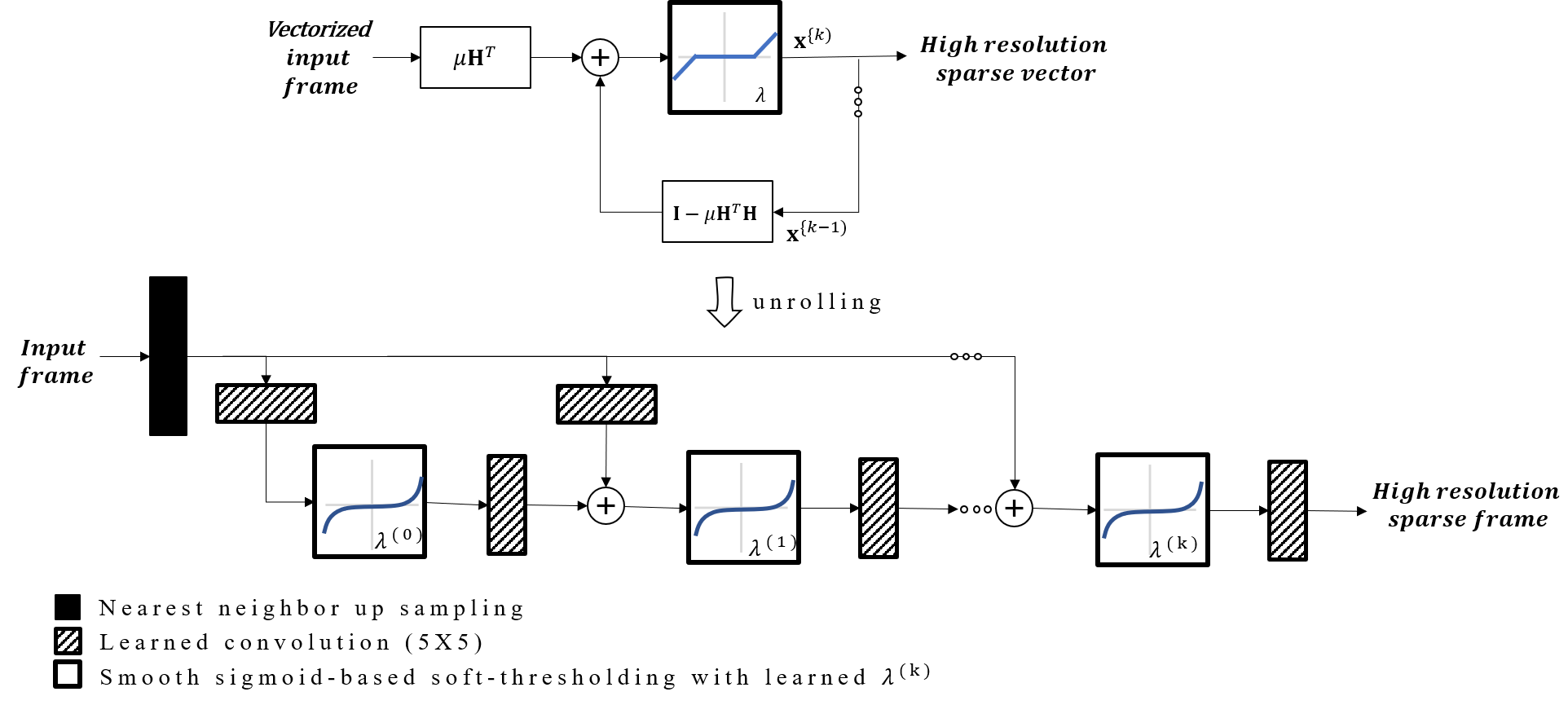}
\caption{Illustration of the deep unrolled ULM architecture. Top: block diagram of the ISTA algorithm; The block with the blue graph is the soft thresholding operator with parameter $\lambda$; The other blocks denote matrix multiplication (from the left side), where $\mu$ is a constant parameter that controls the step size of each iteration and $H$ is a dictionary matrix based on the PSF. Bottom: deep unrolled ULM. Each iteration of the algorithm step is represented as one block of the network. Concatenating these blocks forms a deep neural network. In the unrolled network the parameters of the iterative algorithm are substituted with trainable parameters and convolutional filters. The network is trained through back propagation.} \label{duulm}
\end{figure}
The following procedure was implemented in Python using the Tensorflow framework on a desktop with a 3.59 GHz AMD Ryzen 7 3700X 8-core processor and a NVIDIA GeForce GTX 1650 graphics card.

Detection was performed by applying deep unfolded ULM that makes effective use of sparsity to promote localization with dense concentration of UCAs. UCAs distribution within a frame is highly sparse on a high-resolution grid~\cite{van2017sparsity}. This implies that the support of the resolved image contains only a few non-zero values, allowing to pose the localization task as a sparse image recovery problem~\cite{bar2018sushi,solomon2019exploiting}. The architecture of the network, illustrated in Fig.~\ref{duulm}, is devised by unrolling the Iterative Soft Threshold Algorithm (ISTA)~\cite{daubechies2004iterative}, a popular method for sparse recovery, into a deep neural network where the parameters of the algorithm are replaced with learnable parameters and convolutional filters. Since each layer of the network assimilates an action from the algorithm, the unrolled network naturally inherits prior structures and domain knowledge, rather than learn them from intensive training data, prompting its generalization ability.

Training is performed in a supervised manner. To overcome the lack of sufficient amount of available data, the network is trained using on-line synthesized data of corresponding low-resolution inputs and high resolution targets. Specifically, the targets contain the most basic primitives of the CEUS image, point sources, on a high resolution grid, while the diffraction-limited input consists of point sources convolved with a PSF on a low resolution grid. The density, locations, intensities, background noise, and PSF parameters are randomly sampled from distributions that are defined by the user. Thus, robust inference under a wide variety of imaging conditions is achieved. 
The network is trained through loss minimization via the following loss function:
\begin{equation}
    L(\mathbf{X},\mathbf{Y}\mid\theta)=\|f(\mathbf{X}\mid\theta)-\mathbf{G}*\mathbf{Y}\|_2^2+\lambda\|f(\mathbf{X}\mid\theta)\|_1^1
\end{equation}

\noindent where $\mathbf{Y}$ is the target image containing the true UCA locations, $\mathbf{X}$ is the low resolution input, and $f(\mathbf{X}\mid\theta)$ is the network output. Here $\lambda$ (set to 0.01) is a regularization parameter that promotes sparsity of the recovered image, and $\mathbf{G}$ is a Gaussian filter (standard deviation was set to 1 pixel). The use of $\mathbf{G}$ enables the network to be more forgiving to small errors in the localization and promotes convergence of the network.

We used a 9 block deep network, each block consists of the operations applied in one iteration of ISTA as shown in Fig.\ref{duulm}. The network was trained with ADAM optimizer ($\beta_1=0.9$, $\beta_2=0.999$, and an initial learning rate of $5e-4$)~\cite{kingma2014adam}. The batch size and number of epochs were set to 64 and 1000 respectively.

\section{Results}
The cross correlation revealed different out-of-plane motions across the scans. Consequently, subsequences of different sizes were formed. For consistency, from each scan, we chose a subsequence of 1000 frames for further evaluation. Each of the chosen sequences were aligned and filtered to extract moving UCAs signals. Next, the filtered data was used as input to the network in order to recover the super resolved image.

\begin{figure}
\includegraphics[width=\textwidth]{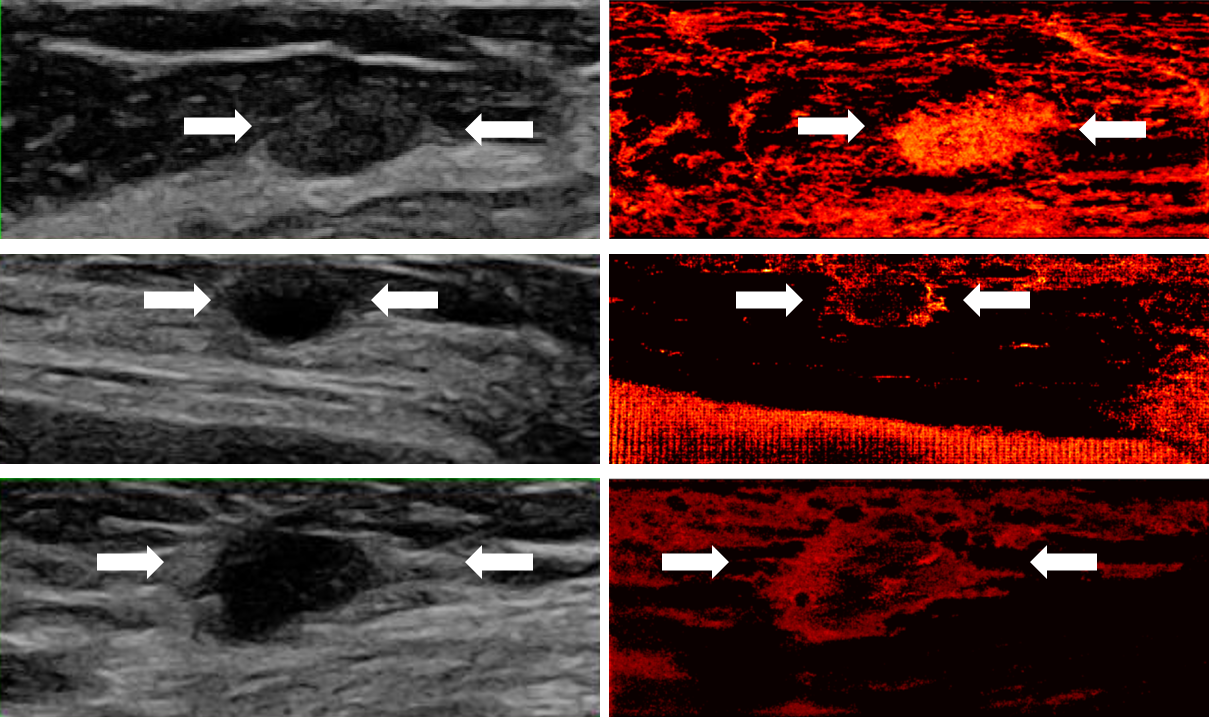}
\caption{Super resolution demonstrations in human scans of three lesions in breasts of three patients. Left: B-mode images. Right: super resolution recoveries. The white arrows point at the lesions; Top: fibroadenoma (benign). The super resolution recovery shows an oval, well circumscribed mass with homogeneous high vascularization. Middle: cyst (benign). The super resolution recovery shows a round structure with high concentration of blood vessels at the periphery of the lesion. Bottom: invasive ductal carcinoma (malignant). The super resolution recovery shows an  irregular mass with ill-defined margins, high concentration of blood vessels at the periphery of the mass, and a low concentration of blood vessels at the center of the mass.}\label{sr1} 
\end{figure}

Fig.~\ref{sr1} presents the super resolution recoveries. The top row displays a fibroadenoma, the middle row displays a cyst, and the bottom row displays an invasive ductal carcinoma. The super resolution recoveries are shown together with the corresponding B-mode (standard) scans. The fibroadenoma recovery depicts an oval, well circumscribed mass with homogeneous high vascularization, the cyst displays a round structure with high concentration of blood vessels at the periphery of the lesion, and the malignant mass recovery displays irregular mass with ill-defined margins, high concentration of blood vessels at the periphery of the mass, and a hypoechoic region at the center of the mass. The correspondence between the structural characteristics of the recovered vasculature of the different lesions and their histological known structure was authenticated by dedicated breast radiologists.

\begin{figure}
\includegraphics[width=\textwidth]{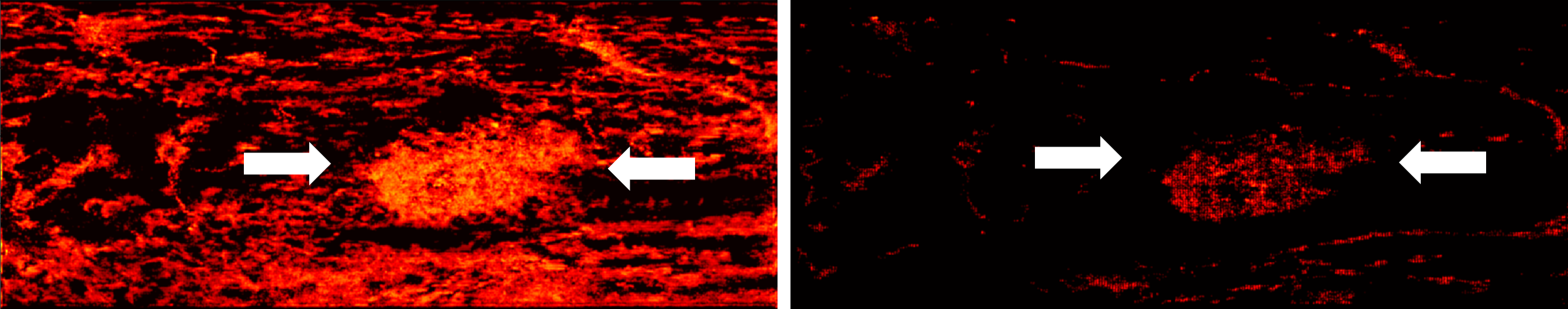}
\caption{Super resolution reconstructions of (the same) in human scan of fibroadenoma. Left: deep unrolled ULM (ours). Right: classical ULM.}\label{sr2} 
\end{figure}

Fig.~\ref{sr2} compares between two super resolved images of the fibroadenoma. The left image was recovered with our method, while the right image was recovered with the classical ULM technique in which local maxima are computed for each frame. Since the computation is applied on the preprocessed data (as described in the previous section), the detected maxima are assumed to correspond to UCAs. Then, the intensity-weighted center of mass of each UCA signal is calculated to obtain coordinates of its localized position. Finally, all localizations are summed to obtain the final image. While both methods reveal similar patterns, a higher density of UCA localizations is observed in the left image via deep unrolled ULM.

\section{Discussion}
This work attempts to enhance the diagnosis and monitoring of breast cancer using US imaging, in order to enable faster, safer, and more accessible treatment using a non-ionizing US device. We demonstrated the feasibility of implementing super resolution US in vivo in humans using a clinical US scanner and standard clinical procedure of contrast administration, promoting simple assimilation of the methods into clinical practice. We showed super resolution recoveries of three different breast lesions. The results successfully demonstrated different morphological features of the various lesions, assisting the differentiation between them. In clinical practice, the physicians distinguish between the lesions according to different characteristics, such as the lesion boundaries and its echogenicity~\cite{gokhale2009ultrasound}. Identifying these differences might be hard for the untrained eye as can be seen by viewing the B-mode images (Fig.~\ref{sr1}, left column). Exploring the super resolution recoveries (Fig.~\ref{sr1}, right column) reveals a different vascular profile for each lesion and improves lesion characterization using US. Still, the lack of ground truth remains a challenge. To cope with this challenge, we consulted with experts to confirm the histological reasoning of the microvascular profiles that were recovered. Furthermore, a non-learning-based method was used as reference. Both methods showed similar profiles further supporting the findings. Nonetheless, exploring Fig.~\ref{sr2} reveals that denser UCA localizations are achieved via the learning-based method (ours) which promotes recoveries in highly populated regions.    

Super-resolution US imaging that leverages deep learning for clinical applications is an exciting opportunity, enabling to harness state of the art technology for noninvasive, robust, precise, and fast in-vivo microvascular imaging. Yet, in clinical application, the black box nature of deep neural networks can hinder the trust of experts in the recoveries. The technique of unrolling~\cite{monga2021algorithm}, used to devise the network architecture, facilitates a highly interpretable network whose reasoning can be easily reflected to the physician. Hence, allowing the physician a better control of the diagnostic process and a means to understand the origin of the artifacts if introduced by the algorithm and identify potential failure cases. While the use of deep learning enables to devise a tool that is fast and has a high generalization ability, the lack of a sufficient amount of in vivo data imposes a great challenge when training. To overcome this difficulty we synthesized data of the CEUS image basic building blocks. The synthesized data enables access to an abundant amount of data required for training, without dependency on prior in vivo data, while training on the image basic building blocks helps the network to avoid overfitting to specific vascular structures.

When tuning the pixel size, the clinical requirements must be considered. In this work, the recovery was performed using a high resolution grid with a 31.25 µm axial/lateral pixel resolution. This enabled the visualization of blood vessels with a resolution under 100 µm, corresponding to venules and arterioles. Recovery at the chosen resolution enabled to address the clinical need of increasing specificity of the lesions.  

We hope that this research will contribute to the ongoing efforts for breast cancer early detection and treatment. The impact of this work can go beyond the field of breast imaging by enhancing the treatment capabilities of vasculature effected pathologies (e.g., enhancing the differentiation between inflammatory and fibrotic phases of disease, and thus effecting treatment choices) by using the proposed technology that makes effective use of clinical US scanners and an interpretable artificial intelligence framework.

\subsubsection{Acknowledgements}
The authors acknowledge the contribution of prof. Ruud van Sloun to the development of the algorithm used in this work.
%
%
%
\bibliographystyle{splncs04}
\bibliography{paper1786_bib}

\end{document}